# Collaborative Feature Aggregation for Face Super-Resolution and Robust Re-Identification

Juheon Hwang[1], Taewan Kim[2], Jiwoo Kang[3*]

[1]Department of Electrical and Electronic Engineering, Yonsei University, Seoul, 03722, South Korea.
[2]Data Science Major, Dongduk Women's University, Seoul, 02748, South Korea.
[3*]Division of Artificial Intelligence Engineering, Sookmyung Women's University, Seoul, 04310, South Korea.

*Corresponding author(s). E-mail(s): jwkang@sookmyung.com;
Contributing authors: consecrated.h@yonsei.ac.kr;
kimtwan21@dongduk.ac.kr;

**Abstract**

We propose a novel collaborative approach for face super-resolution (SR) and robust person re-identification from sequential or multi-view facial images. Traditional SR methods often suffer from blurring and distortion in faces recovered from poor-quality images due to low resolution. Image- and video-based facial SR methods using facial landmarks or segmentation also have similar challenges. To overcome these limitations, we leverage multiple correlated facial observations, across time or viewpoints, by introducing a transformer-based collaborative feature aggregation method that unifies identity features from multi-sequence or multi-view data. This allows faces in multiple sequences of an individual to contribute to accurately estimating common facial features. Furthermore, we propose a cascade SR network to progressively restore the high-resolution image of the target's face with gradual facial feature unification. The unified identity representation is further utilized in person re-identification scenarios, enabling accurate matching even under severe image degradation. The exhaustive experimental results and comparisons show that our method outperforms other state-of-the-art methods, demonstrating consistent improvements in both face super-resolution and re-identification performance . Our work highlights the effectiveness of joint identity reconstruction and progressive image restoration from multiple facial inputs in enhancing downstream visual recognition tasks.

# 1 Introduction

Image super-resolution (SR) is a research field in computer vision that generates high-resolution images from low-resolution images. With the development of deep learning, lots of research has been conducted on SR [1–3], which has been applied to many industries, such as crime prevention systems for surveillance. The surveillance system [4, 5] uses video recording and storage devices, such as CCTV, to identify people involved when an incident occurs, and mainly requires computer vision tasks such as re-identification, face classification, recognition, and face SR. Especially, it is required to verify the identity of the person in order to identify a specific person. Thus, accurate capture of facial images is one of the important technologies in computer vision.

Previous SR methods [6, 7] mainly extract image features from low-resolution input images and use them to create high-resolution images. Since these methods use only the features of the low-resolution image, there is a limitation that they cannot properly restore the high-resolution face in the image when there is insufficient information due to the degradation of the low-resolution image.

Recently, some SR methods [8, 9] have used implicit neural representations based on 2D images, such as neural radiance fields commonly used in 3D domains. These methods use an implicit neural representation, so they can perform the SR task more reliably by learning continuous representation of the target in the image coordinates. However, existing methods, including the state-of-the-art method (*e.g.,*IDM) [9], suffer from blurring and distortion when performing SR on highly low-resolution images, as described in Fig. 1 (b).

Some efforts have been made to deal with details that cannot be restored by low-resolution images alone. Several methods [11–14] have used prior information on faces, such as facial landmarks or facial segmentation, for face SR. However, these methods rely heavily on the learned prior knowledge to overcome the lack of information in low-resolution images, so the SR results are prone to differ significantly from the real ones. In other studies, video sequences are used as input [15–18] for face SR. However, these methods do not show significant improvement over existing work on single-image face SR or video SR because they do not sufficiently account for the correlation between sequences.

Therefore, we address these issues by introducing a novel collaborative face SR method for sequential or multi-view face images. Specifically, we introduce a transformer-based collaborative identity aggregation to fully utilize face video sequences or multiple images captured from different viewpoints at different times. We aggregate identity features from multiple facial images into a single feature identity indicator by compensating for insufficient information from a single face and use the unified identity prior as a conditional input to the SR network. Furthermore, we propose a cascade network structure for SR to progressively restore the target's facial image and facial identity in detail. Our cascade SR network is designed as a multi-stage restoration process, where each stage predicts the residual (*i.e.,*delta) between the current output and the target high-resolution image. This residual is added to the input of the current stage to refine facial details progressively across stages. This structure helps to reduce blurring by gradually recovering high-frequency facial components while maintaining stability in low-quality inputs. For collaborative identity

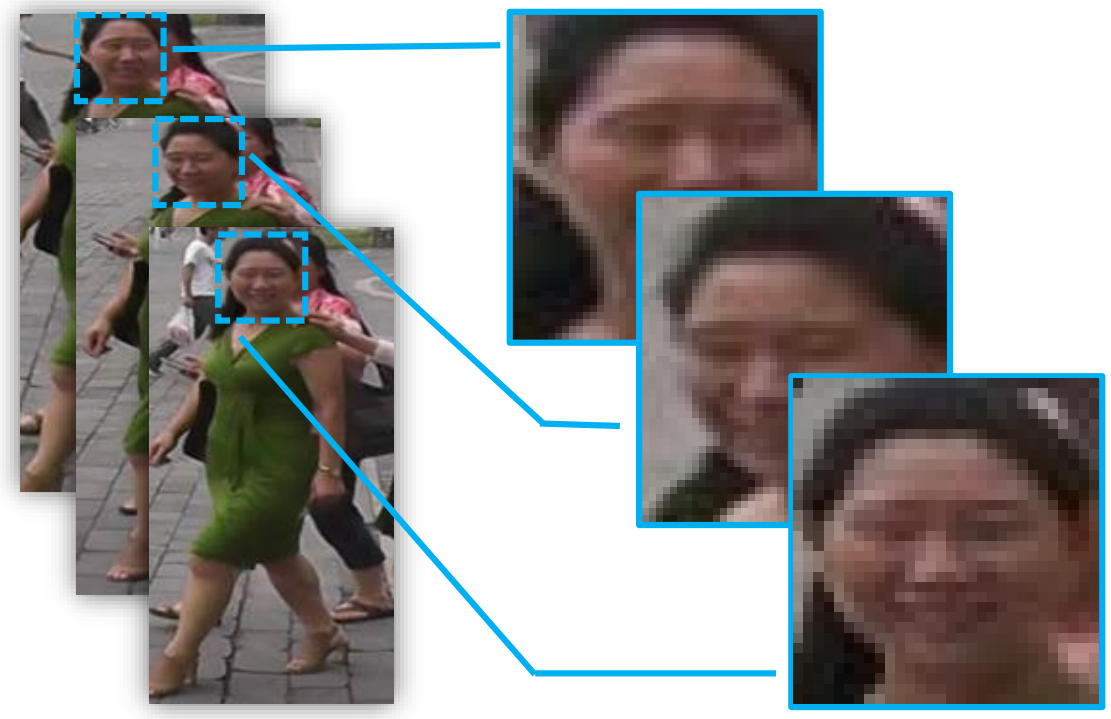

(a) Low-resolution facial surveillance sequences of the same identity.

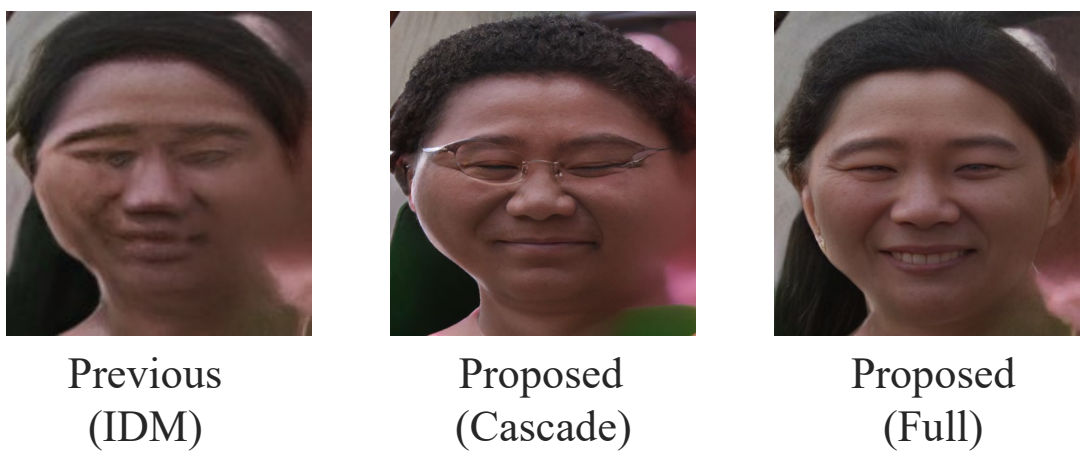

(b) Super-resolution results with a state-of-the-art method (IDM), the proposed cascade superresolution network without identity collaboration, and the proposed cascade superresolution network with collaborative identity reconstruction, respectively.

**Fig. 1**: Face super-resolution examples from low-resolution images. The proposed method generates high-quality images from low-resolution images with a cascade super-resolution network and collaborative identity reconstruction. Face regions are extracted from images in surveillance datasets, MARS [10].

reconstruction, we extract facial features from multiple input images using a shared identity encoder and then apply a transformer-based self-attention module to model inter-image correlations. The resulting unified identity embedding captures consistent identity-related semantics across variations in pose, illumination, or resolution. This aggregated identity prior is injected into each cascade stage to guide the restoration with consistent identity-aware information. Thus, the facial image and its facial identity are restored accurately by gradually adding facial details. The experimental results show that the proposed SR method achieved significant improvements over the state-of-the-art methods. We also perform experiments for re-identification scenarios by using the common feature embedding of multiple images. The results demonstrate significant advantages of the proposed method in SR by efficiently recovering the shared information across images.

Our contributions are summarized as follows:

- We propose a novel collaborative face SR method that efficiently uses several facial features from multiple images with cascade architecture to restore a high-quality SR image from noisy and low-quality images.
- We introduce a transformer-based collaborative aggregation network to estimate common facial features in multiple images of an individual.
- We propose a cascade SR network to gradually restore the accurate target's facial images and facial features with high-resolution details.
- Through experiments, it is verified that the proposed method outperforms the state-of-the-art SR methods, demonstrating the strong benefits of the proposed methods in surveillance and reconnaissance.

# 2 Related Works

## 2.1 Image Super-resolution

Image super-resolution methods are generally divided into regression-based and generative model-based methods. Regression-based methods, such as SwinIR [6], LIIF [8], and VQFR [14], learn mappings to generate high-resolution images from low-resolution images. Specifically, LIIF [8] used implicit representation to replace the traditional convolutional network with a multilayer perceptron, allowing reliable SR through continuous representation in 2D. The generative model-based method, such as ESRGAN [19], Real-ESRGAN [7], and IDM [9], used generative models to generate high-resolution images using low-resolution images as input or condition. Specifically, IDM [9] restored a high-resolution image from noise using a flow-based method with a low-resolution image as a condition. However, these methods mainly use low-resolution images to generate high-resolution images, and thus, they are fully dependent on the learned information, so they are limited in their ability to generate high-quality faces from low-quality input. Considering the limitations of a single source image and the nature of the surveillance environment, we incorporate common facial identities in multiple facial images, through a collaborative identity network proposed to obtain accurate and high-quality facial images.

## 2.2 Implicit Neural Representation

Implicit Neural Representation has been used in 3D coordinate-based rendering methods. It has shown excellent performance in 3D shape or surface learning [20, 21] and 3D spatial volume rendering [22]. Neural radiance fields (NeRF) [22], a pioneering work of implicit neural representation, has introduced a method to learn 3D scenes by predicting colors in 3D coordinates using multi-layer perceptrons. Due to the power of implicit neural representation to represent 3D scenes, several efforts have been made to extend it to 2D images. LIIF [8] used 2D local coordinates in SR tasks to predict the pixel color of a desired location from an image feature latent vector. IDM [9] used denoising diffusion probabilistic models (DDPM) [23] to condition an input low-resolution image for a denoising network so that the network restores the image from noise according to the condition. However, as these methods use 2D coordinate-based neural implicit representations for features in low-resolution images, their results are prone to blurring. Our method addresses the limitations of these networks by using facial identity features that can be obtained from multiple input images in a collaborative manner.

### 2.3 Person Re-identification

Person re-identification is the task of extracting a person's identity from a given image or video and matching it to those captured by other cameras or locations. Along with the progress in image-based re-identification [24–31], there has been a lot of recent progress in video-based re-identification [32–36]. The goal of video-based re-identification approaches is to generate a person representation that utilizes spatial and temporal information robust to various factors such as person location and occlusions. To leverage the image-level features in sequence-level representations, a temporal memory module has been introduced in [37] to capture attention optimized for common patterns of sequences. By exploring intra-frame spatial correlations to determine the attentional weight of each location, and temporal coherence information, the work in [38] suppressed interfering features and enhanced discriminative power. On the other hand, the work in [39] used a flexible unit for retrieving complementary information along spatiotemporal dimensions. For face SR, our method extracts the shared identity of the target person through a collaborative identity network. Through the collaborative strategy, the shared information of multiple images is embedded in the encoder of the cascade restoration model. Therefore, we perform experiments for re-identification scenarios by using the shared feature embedding of multiple images. Through experiments and comparisons on re-identification scenarios, it is demonstrated that our proposed method can accurately extract common information from sequences.

## 3 Method

Our method assumes $N$ low-resolution facial images of an individual. To obtain the target facial identity information from multiple facial images, we propose a transformer-based collaborative identity network to extract a unified facial identity, as depicted in Fig. 2 (a). Once the unified facial identity is predicted from multiple facial images, the cascade restoration model uses the facial identity as prior information to perform the restoration of high-quality images from low-resolution facial images, as shown in Fig. 2 (b).

Finally, in our novel cascade SR method with collaborative identity features, we utilize the unified facial identity obtained from multiple facial images through the collaborative identity network and iterate the cascade network $M$ times to obtain the final result, as depicted in Fig. 2 (c). In the following, we first introduce the collaborative identity network and then the cascade SR framework in detail. Subsequently, the loss functions defined for our framework are presented.

### 3.1 Collaborative Identity Network

Our face SR network uses the facial identity features extracted via the facial identity network as the condition input to restore the facial image. Due to different facial poses

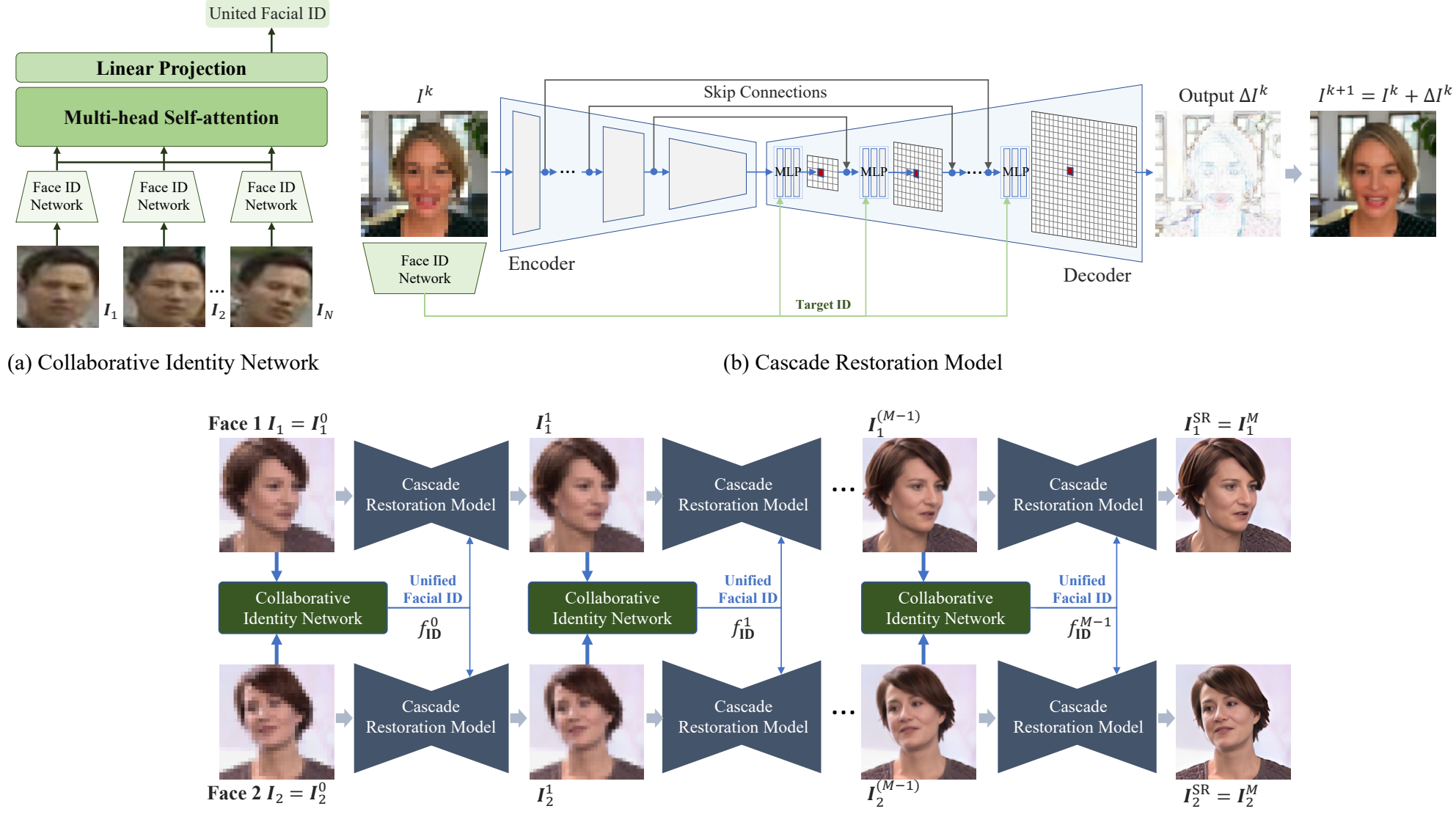

(a) Collaborative Identity Network (b) Cascade Restoration Model

(c) Cascade Superresolution Framework with Collaborative Identity ($N$=2 case)

**Fig. 2**: The overview architectures of our cascade super-resolution framework with collaborative identity from $N$ images with $M$ collaborative iterations. (a) Collaborative identity network based on a transformer architecture. (b) Cascade restoration network with UNet architecture that uses facial identity to complement the image feature. (c) The proposed cascade SR network with collaborative identity from multiple facial images of an individual. Here, the case of collaborative supperresolution using two facial images of the same person is depicted ($N = 2$).

and the lower resolution of the input images, the facial identity features extracted from low-resolution images can be significantly different, even if the face belongs to the same person. To address this problem, we aggregate facial features in multiple images to accurately obtain a unified facial identity with a transformer-based collaborative identity network, as described in Fig. 2 (a).

Assume that $N$ facial images of an individual are given as input, *i.e.*,$(\boldsymbol{I}_1, \boldsymbol{I}_2, ..., \boldsymbol{I}_N)$. The unified facial identity $f_{\text{ID}}$ is obtained as follows:

$$f_{\text{ID}} = \text{Attn}(\phi(\boldsymbol{I}_1), \phi(\boldsymbol{I}_2), ..., \phi(\boldsymbol{I}_N)) \tag{1}$$

where $\text{Attn}(\cdot)$ is the multi-head self-attention network [40], and $\phi(\cdot)$ is the facial identity feature network. We used a state-of-the-art face recognition model [41] for the identity feature network. We note that our method does not distinguish any primary or secondary image among the $N$ input images. All images are treated equally within the attention-based identity aggregation. To avoid any potential ordering bias, the order of input images is randomly shuffled during training. This ensures that the model remains invariant to the input sequence and focuses solely on the semantic content of each image.

### 3.2 Cascade Restoration Model

The input and output image sizes of the cascade network are equivalent because the cascade restoration network operates in a cascade manner. Thus, linear interpolation is used for upsampling the input image to the desired SR size in advance, and then the restoration network performs the SR by iteratively finding the difference, or the delta, toward a high-quality face.

Our cascade restoration network is made up of an encoder and a decoder. Each layer in the encoder has a skip connection to the corresponding layer in the decoder, as described in Fig. 2 (b). The skip connection helps our cascade restoration network to preserve the details in the original image and learn the hierarchical facial features. Specifically, the convolutional neural network-based encoder generates latent codes of low-resolution image features. The decoder finds the facial details from the latent codes using MLP-based image implicit neural representations [8]. The decoder restores the SR image by utilizing a facial identity prior extracted from the low-resolution input image. The identity prior helps the decoder to produce high-quality faces by providing facial information, such as face appearance or the shapes of eyes, nose, and mouth.

Assume that $N$ different low-resolution facial images of an individual are given. For the $i^{th}$ facial image ($1 \leq i \leq N$), the cascade restoration model iteratively finds the delta RGB values from the input low-resolution image of the $i^{th}$ face and the facial identity feature as follows:

$$\boldsymbol{I}_i^{k+1} = \mathrm{CRM}\left(\boldsymbol{I}_i^k, \phi(\boldsymbol{I}_i)\right) = \boldsymbol{I}_i^k + \Delta \boldsymbol{I}_i^k \tag{2}$$

where $\mathrm{CRM}(\cdot)$ is the cascade restoration model, $\Delta \boldsymbol{I}_i^k$ is the delta difference estimated in the $k^{th}$ iteration ($1 \leq k \leq M$), $\boldsymbol{I}_i$ is the initial low-resolution image, and $\boldsymbol{I}_i^0 = \boldsymbol{I}_i$. We use the cascade restoration model recursively $M$ times in a cascade manner, producing the final SR image, $\boldsymbol{I}_i^{\mathrm{SR}} = \boldsymbol{I}_i^M$ of the $i^{th}$ image. In our experiments, we empirically set the cascade iteration to $M = 3$.

### 3.3 Cascade SR with Collaborative Identity

The cascade restoration network utilizes the facial identity information of each input face for SR. In other words, it uses the facial identity features of the $i^{th}$ image to restore the $i^{th}$ face. If we have multiple images of an individual, the common identity of the multiple images can be used for SR. Thus, for given $N$ facial images, the formulation in (2) is represented in terms of the unified facial identity, $f_{\mathrm{ID}}$ in (1), by jointly utilizing all input faces for the SR as follows:

$$\boldsymbol{I}_i^{k+1} = \mathrm{CRM}\left(\boldsymbol{I}_i^k, f_{\mathrm{ID}}\right). \tag{3}$$

In addition, the unified identity in (1) can be obtained for each cascade iteration $k$, not only for the initial iteration, as follows:

$$f_{\mathrm{ID}}^k = \mathrm{Attn}(\phi(\boldsymbol{I}_1^k), \phi(\boldsymbol{I}_2^k), ..., \phi(\boldsymbol{I}_N^k)). \tag{4}$$

Thus, we further extend the cascade restoration model in (3) to use the unified identity obtained at each cascade iteration. Finally, our cascade SR framework with collaborative identity in Fig. 2 is formulated as follows:

$$\boldsymbol{I}_i^{k+1} = \mathrm{CRM}\left(\boldsymbol{I}_i^k, f_{\mathrm{ID}}^k\right). \tag{5}$$

### 3.4 Loss Functions

This section describes the loss functions used in our face SR method. We train our proposed face SR framework using image loss $\mathcal{L}_{\mathrm{img}}$ and identity loss $\mathcal{L}_{\mathrm{ID}}$ to recover a high-resolution image and identity for the $i^{th}$ face from multiple low-resolution facial images as follows:

$$\mathcal{L}_{\mathrm{loss}}(\boldsymbol{I}_i^{\mathrm{GT}}, \boldsymbol{I}_i^{\mathrm{SR}}) = \mathcal{L}_{\mathrm{img}}(\boldsymbol{I}_i^{\mathrm{GT}}, \boldsymbol{I}_i^{\mathrm{SR}}) + \lambda_{\mathrm{ID}}\mathcal{L}_{\mathrm{ID}}(\boldsymbol{I}_i^{\mathrm{GT}}, \boldsymbol{I}_i^{\mathrm{SR}}), \tag{6}$$

Thus, the overall loss function $\mathcal{L}_{\mathrm{all}}$ for our framework is defined by accumulating the loss for each facial image in (6) over $N$ faces as follows:

$$\mathcal{L}_{\mathrm{all}} = \sum_{i}^{N} \mathcal{L}_{\mathrm{loss}}(\boldsymbol{I}_i^{\mathrm{GT}}, \boldsymbol{I}_i^{\mathrm{SR}}). \tag{7}$$

where $\boldsymbol{I}_i^{\mathrm{GT}}$ and $\boldsymbol{I}_i^{\mathrm{SR}}$ are the ground-truth image and the SR image of the $i^{th}$ face, respectively, and $\lambda_{\mathrm{ID}}$ is the balancing constant between image and identity losses. We set the balancing constant to $\lambda_{\mathrm{ID}} = 1.0$ for our experiments. Note that each low-resolution input image $\boldsymbol{I}_i$ has a corresponding high-resolution ground-truth image $\boldsymbol{I}_i^{\mathrm{GT}}$ in our formulation. The SR image $\boldsymbol{I}_i^{\mathrm{SR}}$ is restored independently for each $i$ using the shared identity prior, and losses are computed per image. As such, Eq. (6) and Eq. (7) represent the per-image loss and the aggregated loss over all $N$ inputs, respectively.

Specifically, the image loss measures the $\mathbb{L}_1$ distance between the RGB values of $\boldsymbol{I}_i^{\mathrm{SR}}$ and $\boldsymbol{I}_i^{\mathrm{GT}}$ to generate the SR image $\boldsymbol{I}_i^{\mathrm{SR}}$. It allows our network to accurately represent the original faces $\boldsymbol{I}_i^{\mathrm{GT}}$ using the low-resolution image features and the identity of the face in a cascade manner. Thus, the image loss is formulated as follows:

$$\mathcal{L}_{\mathrm{img}}(\boldsymbol{I}_i^{\mathrm{SR}}, \boldsymbol{I}_i^{\mathrm{GT}}) = \left\|\boldsymbol{I}_i^{\mathrm{SR}} - \boldsymbol{I}_i^{\mathrm{GT}}\right\|_1. \tag{8}$$

The identity loss measures the $\mathbb{L}_1$ distance between the identity features of the SR image obtained from the network and the ground-truth image as follows:

$$\mathcal{L}_{\mathrm{ID}}(\boldsymbol{I}_i^{\mathrm{SR}}, \boldsymbol{I}_i^{\mathrm{GT}}) = \left\|\phi(\boldsymbol{I}_i^{\mathrm{SR}}) - \phi(\boldsymbol{I}_i^{\mathrm{GT}})\right\|_1. \tag{9}$$

## 4 Experiments

### 4.1 Dataset

In our paper, we assume the multi-face SR for sequential or multi-view images. Therefore, it is difficult to use high-quality facial image datasets, such as FFHQ [42] and CelebA-HQ [43], because they do not provide multiple images of an individual. Additionally, face video datasets, such as the existing VoxCeleb [44–46] are not appropriate to use for our experiments because their resolution and quality are not high enough to train SR networks. Finally, we used VFHQ [47] and CelebV-HQ [48], which are face video datasets that satisfy both temporal and high-quality properties, for our experiments. VFHQ [47] comprises facial clips with a resolution between 700 × 700 and 1000 × 1000. CelebV-HQ [48] has a resolution of 512 × 512, which is higher than VoxCeleb's resolution of 224 × 224 [46].

For the performance comparison, we used the face video dataset and face multi-view dataset to evaluate the performance of the proposed method on sequential and multi-view images, respectively. We used the test sets of VFHQ [47] and CelebV-HQ [48] for the face video dataset, and the multiface dataset [49] for the face multi-view dataset. The multiface dataset [49] consists of sequences of 65 different expressions across 13 unique identities. The number of cameras used to capture the images was between 40 and 160. The source images have a resolution of 2668 × 4096. However, to use them in our comparison experiment, they were resized to preserve their proportions and padded to make the resolution equivalent horizontally and vertically.

### 4.2 Implementation Details

We trained our cascade SR framework in an end-to-end manner with facial images resized to a resolution of 256 × 256. Following standard settings in recent face SR methods [6, 8, 9], we define low-resolution (LR) images as 32×32 images obtained by bicubic downsampling from 256×256 high-resolution (HR) images. This corresponds to an 8× SR scale and ensures fair comparisons across all evaluated methods. We used the Adam optimizer [50] with a fixed learning rate initialized with $1e^{-4}$. The models were trained for 4000 epochs with batch size 4. We used the ArcFace [41] face recognition network for the facial identity feature network in Fig. 2. The *PyTorch* framework [51] with a single *Nvidia 3090* GPU was used to perform experiments.

### 4.3 Performance Comparison

To comprehensively validate the effectiveness and robustness of our proposed framework, we evaluate its performance across diverse scenarios and tasks. This section is organized into six subsections: (1) face SR from sequential images, (2) error analysis via difference maps, (3) SR from multi-view inputs, (4) SR for severely degraded real-world images, (5) person re-identification performance, and (6) SR-assisted re-identification.

These experiments are designed to collectively demonstrate the robustness of our method in a wide range of conditions. Specifically, we show that our framework handles both temporally coherent (sequential) and geometrically diverse (multi-view) inputs, performs well under real-world degradations (*e.g.,*low-resolution surveillance images), and effectively supports downstream tasks like person re-identification. In particular, the re-identification experiment on full-body pedestrian images is designed to verify the general applicability of our super-resolution framework beyond face-specific domains.

The inclusion of ablation studies (Section 4.4) provides deeper insight into the contribution of each architectural component within our framework. We first examine the effect of removing the collaborative identity network and the cascade restoration structure, showing that each component is critical for accurate facial detail recovery and identity consistency. Next, we analyze the sensitivity to the number of input facial images used for identity aggregation, and find that even using two images provides sufficient information for reliable reconstruction.

We also evaluate the effect of using different face recognition networks to extract identity features, comparing ArcFace [41] with two recent alternatives: AdaFace [52] and TopoFR [53]. Although all three models yield similar trends, ArcFace provides the best trade-off between identity consistency and perceptual reconstruction quality. These results validate that both the identity-aware design and the progressive cascade architecture are key to ensuring robust and high-fidelity super-resolution across various input conditions.

### 4.3.1 Super-resolution with Sequential Images

We compare our face super-resolution results qualitatively and quantitatively with state-of-the-art SR methods on datasets of sequential and multi-view images, respectively. To comprehensively verify our method, we consider both general-purpose SR methods (*i.e.,*IDM [9]) and face-specific approaches (*i.e.,*VQFR [14]), as well as recent video and multi-view super-resolution techniques designed for sequential or multi-frame inputs. In particular, we include SAVSR [54], a scale-adaptive video SR method, and MVSR [55], a geometry-aware multi-view SR framework, to align the comparisons more closely with the collaborative nature of our proposed approach. IDM [9] proposes a denoising model from random noise using UNet architecture based on DDPM [23] network to perform SR using features extracted from low-resolution images as conditions. VQFR [14] constructs a codebook for the facial image dataset based on VQ-VAE [56] and uses it as a face prior for face SR. SAVSR [54] performs video super-resolution by learning scale-adaptive kernels to reconstruct high-resolution frames from multiple temporally coherent inputs. It leverages both motion information and dynamic filtering for enhanced fidelity. MVSR [55] synthesizes geometry-aware reference views for multi-view image super-resolution. It explicitly models view transformations and uses learned homography for a more accurate aggregation of information across different viewpoints. We evaluate the proposed method using a pair of facial images and also using a single image for fair comparisons with other methods. Thus, we denote our model that uses a pair of images as 'Ours' and one using a single image as 'Ours-'. All methods are compared using $8\times$ SR with $32\times32 \rightarrow 256\times256$.

We measure SR performances in terms of peak signal-to-noise ratio (PSNR), structural similarity index measure (SSIM) [57], learned perceptual image patch similarity (LPIPS) [58], identity consistency (ID) [59], and frechét inception distance (FID) [60] metrics on VFHQ [47] and CelebV-HQ [48] datasets.

PSNR, SSIM, and FID are well-known metrics used to measure the super-resolution performance. LPIPS measures the perceptual similarity between the intermediate layer's activations of two images for pre-trained networks on large-scale datasets. ID measures the identity consistency between the results of the SR network and the ground-truth image. We calculate the cosine similarity of the identity feature vectors acquired through ArcFace [41], which is a recent face recognition network.

**Table 1**: Quantitative comparisons on sequential images from VFHQ [47] and CelebV-HQ [48] datasets.

| | IDM [9] | VQFR [14] | MVSR [55] | SAVSR [54] | Ours- | Ours |
|---|---|---|---|---|---|---|
| PSNR (↑) | 19.65 | 23.03 | 22.21 | 23.76 | 22.82 | 25.58 |
| SSIM (↑) | 0.4589 | 0.6245 | 0.6032 | 0.6485 | 0.6143 | 0.6977 |
| LPIPS (↓) | 0.4982 | 0.4281 | 0.3975 | 0.4198 | 0.4314 | 0.3458 |
| ID (↑) | 0.602 | 0.685 | 0.637 | 0.659 | 0.674 | 0.792 |
| FID (↓) | 69.24 | 54.62 | 57.18 | 53.37 | 55.21 | 51.42 |

Table 1 shows that IDM performs the worst among the compared methods, followed by MVSR and SAVSR. VQFR and the proposed methods (Ours-, Ours) outperform the others, with Ours showing the best performance across all evaluation metrics. This reflects the importance of accurate identity reconstruction and collaborative feature utilization. IDM [9], which utilizes a diffusion-based restoration process conditioned on low-resolution features, tends to generate smooth outputs but lacks the ability to recover identity-specific facial features. As it does not incorporate any facial priors or multi-frame information, it performs poorly in both perceptual similarity (LPIPS) and identity consistency (ID). VQFR [14] enhances visual quality by injecting prior features retrieved from a pre-trained facial codebook. While this allows for sharp and realistic textures, the substituted features may not match the actual subject, often leading to identity inconsistency, especially when the input is highly degraded. MVSR [55], although designed for multi-view SR, tends to over-smooth facial regions and lacks fine-grained structure preservation. Its geometry-aware reference synthesis helps reduce pixel-level errors (reflected in PSNR), but its identity and perceptual scores are limited due to averaging across views. SAVSR [54], which performs video super-resolution using learned scale-adaptive filters, achieves strong PSNR and SSIM by leveraging temporal information. However, in our face SR setting, it produces visually unnatural facial details with increased artifacts, which hurts its LPIPS and ID performance. This suggests that general video SR models do not transfer well to identity-sensitive tasks without explicit identity conditioning.

The proposed method with a single image input (Ours-) already performs comparably to VQFR, thanks to its ability to directly learn identity-aware priors rather than relying on codebook lookups. Our full model (Ours), which aggregates identity features from multiple facial images, consistently achieves the best results across all metrics, including perceptual quality and identity preservation, highlighting the benefit of transformer-based collaborative identity aggregation and progressive cascade restoration. These findings demonstrate that simply utilizing multiple frames (as in SAVSR or MVSR) is not sufficient; rather, explicitly modeling and unifying identity representations across frames is key to accurate and natural face reconstruction. Our method leverages this principle to recover facial details faithfully, even from severely degraded inputs.

The results of the qualitative comparison on sequential images of VFHQ [47] and CelebV-HQ [48] datasets are shown in Fig. 3. IDM generates globally smooth but overly simplified facial regions, failing to restore high-frequency details and identity-specific features. VQFR produces visually appealing results with sharp textures by using a pre-trained codebook, but it often introduces inconsistencies in facial structure, such as misaligned eyes or distorted expressions, due to mismatches between the selected prior and the true identity. MVSR restores facial images with moderate quality, but tends to produce blurred and overly smoothed features across different frames, resulting in loss of facial detail and structure. SAVSR shows relatively high pixel fidelity and sharpness due to its temporal-aware filtering, but the generated faces often exhibit unnatural textures or asymmetric artifacts, especially around the eyes and mouth. Ours- shows better structural consistency and identity alignment than prior-based methods by learning identity priors from a single image. Finally, our full model (*i.e.,*Ours) shows the most visually coherent results, maintaining both global structure and local facial details across all frames. This is due to its collaborative identity network, which effectively aggregates identity features from multiple input frames and guides progressive restoration in the cascade network.

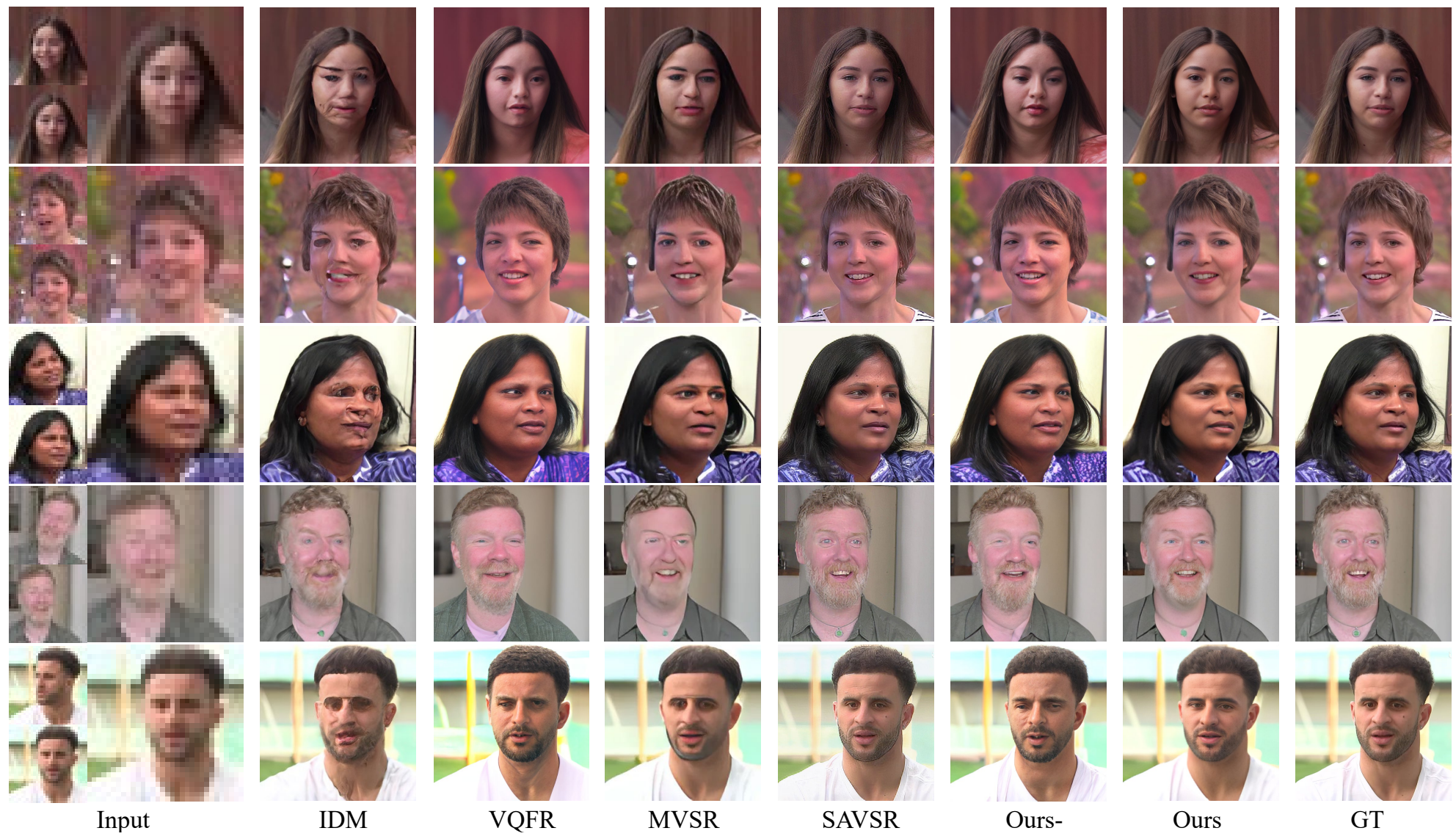

**Fig. 3**: Results of 8x SR on the VFHQ [47] and CelebV-HQ [48] dataset. Our framework produces detailed features on faces most similar to ground-truth faces.

### 4.3.2 Error Analysis

We performed additional experiments on sequential datasets, *i.e.,*VFHQ [47] and CelebV-HQ [48] datasets. The qualitative results of IDM, VQFR, MVSR, SAVSR, and ours on the sequential datasets are illustrated in Fig. 4.

Here, the error map, the difference from the ground truth, is visualized to deeply investigate the results. This error map is the absolute difference between the output of each method and the ground truth.

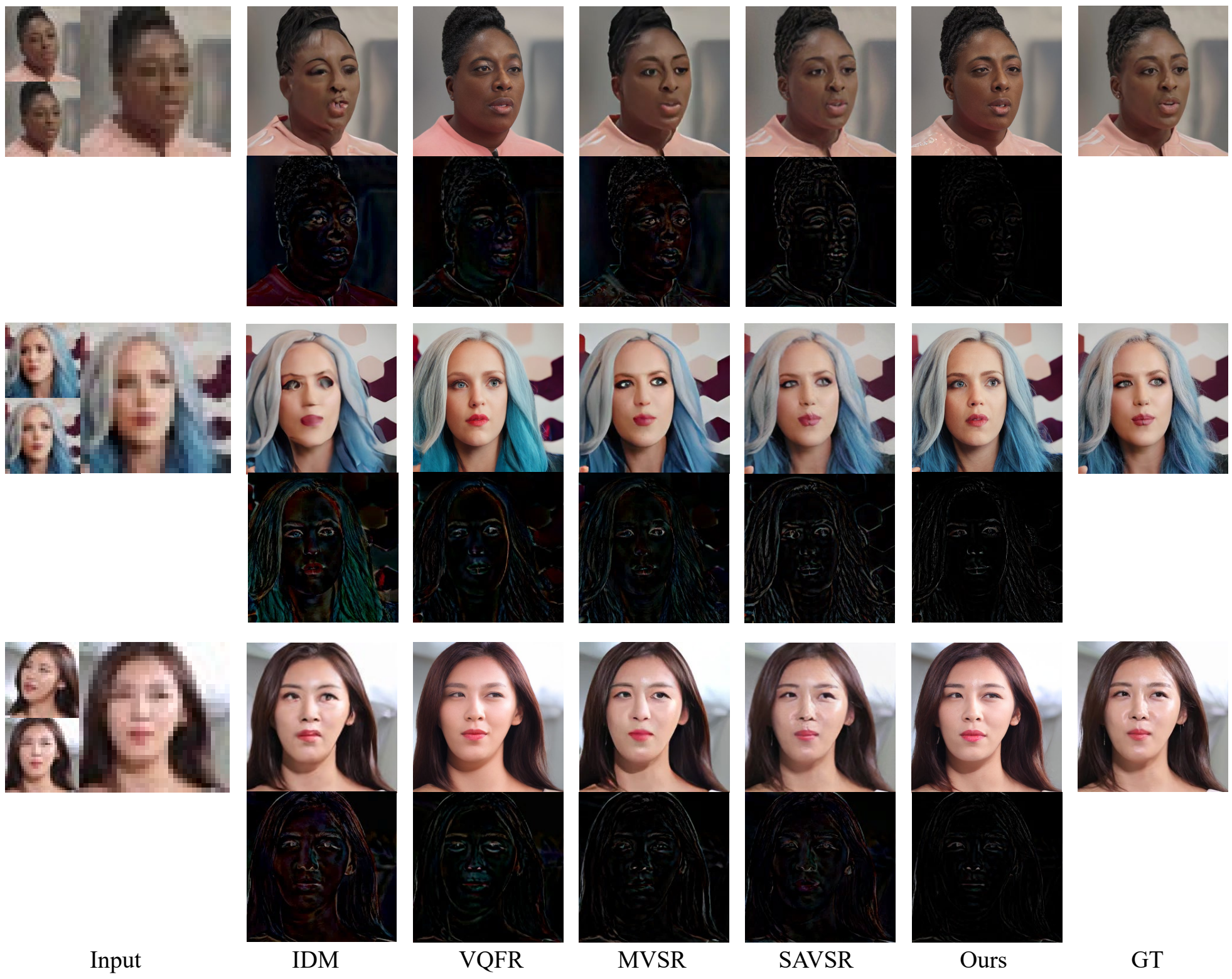

**Fig. 4**: Results of 8x SR on the VFHQ [47] and CelebV-HQ [48] dataset with the error map.

Due to the limited detail recovery capacity of IDM, error maps show large discrepancies along key facial regions such as the eyes, nose, and mouth, indicating poor restoration of identity-specific features. VQFR produces sharper results, but exhibits noticeable errors in regions with complex textures, such as hair, facial contours, and clothing due to mismatches between the codebook features and the actual input. MVSR, which synthesizes views across multiple angles, shows smoother reconstructions but results in consistent blur and loss of definition, particularly at facial boundaries and high-frequency regions, as seen in the error maps. SAVSR, though temporally aware, introduces artifacts in the form of local distortions or asymmetric textures, which appear as sharp residuals in the error maps around the eyes and lips. In contrast, the proposed method shows significantly reduced errors across the entire face, especially in detailed regions, highlighting its ability to utilize shared information across frames to recover identity-preserving and perceptually accurate structures.

#### 4.3.3 Super-resolution with Multi-view Images

To further evaluate the generalizability and robustness of the proposed method under heterogeneous input conditions, we conduct qualitative and quantitative experiments on multi-view facial images. In this setting, we compare our approach with four representative super-resolution methods: IDM [9], VQFR [14], MVSR [55], and SAVSR [54]. These methods represent various architectural paradigms, including diffusion-based image generation (IDM), prior-based facial restoration (VQFR), geometry-aware multi-view reconstruction (MVSR), and scale-adaptive video SR (SAVSR). All models are evaluated using an 8× SR setup, transforming 32×32 low-resolution inputs into 256×256 high-resolution outputs. Quantitative comparisons using PSNR, SSIM, LPIPS, ID, and FID metrics on the multiface dataset are summarized in Table 2.

**Table 2**: Quantitative comparisons on multi-view images from the multiface dataset[49].

| | IDM [9] | VQFR [14] | MVSR [55] | SAVSR [54] | Ours |
|---|---|---|---|---|---|
| PSNR (↑) | 16.28 | 18.87 | 20.92 | 20.21 | 22.54 |
| SSIM (↑) | 0.4278 | 0.5738 | 0.6145 | 0.5932 | 0.6489 |
| LPIPS (↓) | 0.5126 | 0.4571 | 0.4048 | 0.4267 | 0.3725 |
| ID (↑) | 0.553 | 0.627 | 0.684 | 0.662 | 0.743 |
| FID (↓) | 74.89 | 63.14 | 58.77 | 60.11 | 53.08 |

As shown in Table 2, the proposed method outperforms all comparison methods across all evaluation metrics. IDM and VQFR show the lowest performance overall, consistent with their single-image design and lack of cross-view identity modeling. MVSR demonstrates improved performance over VQFR and IDM, particularly in structural metrics such as SSIM and ID, benefiting from its geometry-aware multi-view synthesis. However, its outputs tend to be overly smoothed, limiting perceptual quality and fine detail recovery. SAVSR achieves competitive pixel-wise performance (*e.g.,*PSNR), but shows inferior identity and perceptual metrics (ID, LPIPS) due to temporal filtering artifacts and its limited ability to adapt to spatially diverse multi-view inputs. In contrast, our method leverages a transformer-based collaborative identity network to robustly aggregate identity features across views, regardless of pose or input order. Combined with the cascade restoration model, this enables accurate reconstruction of identity-preserving high-resolution images. These results demonstrate that explicitly modeling shared identity representations across multiple views, rather than simply combining pixel-level features, is the key to achieving high-quality and identity-consistent face super-resolution in multi-view settings.

Qualitative performance comparisons on multi-view dataset are illustrated in Fig. 5. IDM [9], which applies a denoising diffusion model conditioned on low-resolution input, produces outputs that are globally smooth but lacking in semantic detail. As shown in the figure, key facial components such as the eyes, mouth, and nose remain blurry and ill-defined, and the restored faces often appear generic rather than identity-specific. This is due to IDM's limited capacity to recover high-frequency features without explicit identity guidance or multi-frame context. VQFR generates visually sharp images using features from a pre-trained VQ-VAE codebook, but often introduces spatial distortions in key components such as the eyes and nose, as the retrieved features may not fully match the target identity. MVSR, while leveraging multi-view geometry, tends to average over viewpoints and results in overly smoothed textures. Although structural alignment is relatively preserved, the outputs lack fine-grained detail and appear perceptually flat. SAVSR, which is designed for video SR, does not generalize well to multi-view settings with high spatial variation. It introduces localized artifacts and noise, particularly in facial regions, due to the lack of view-specific identity conditioning. In contrast, our method restores both global structure and local textures more accurately by aggregating identity features across views through the collaborative identity network. As a result, our outputs maintain sharper facial components and better preserve the identity and expression of the subject.

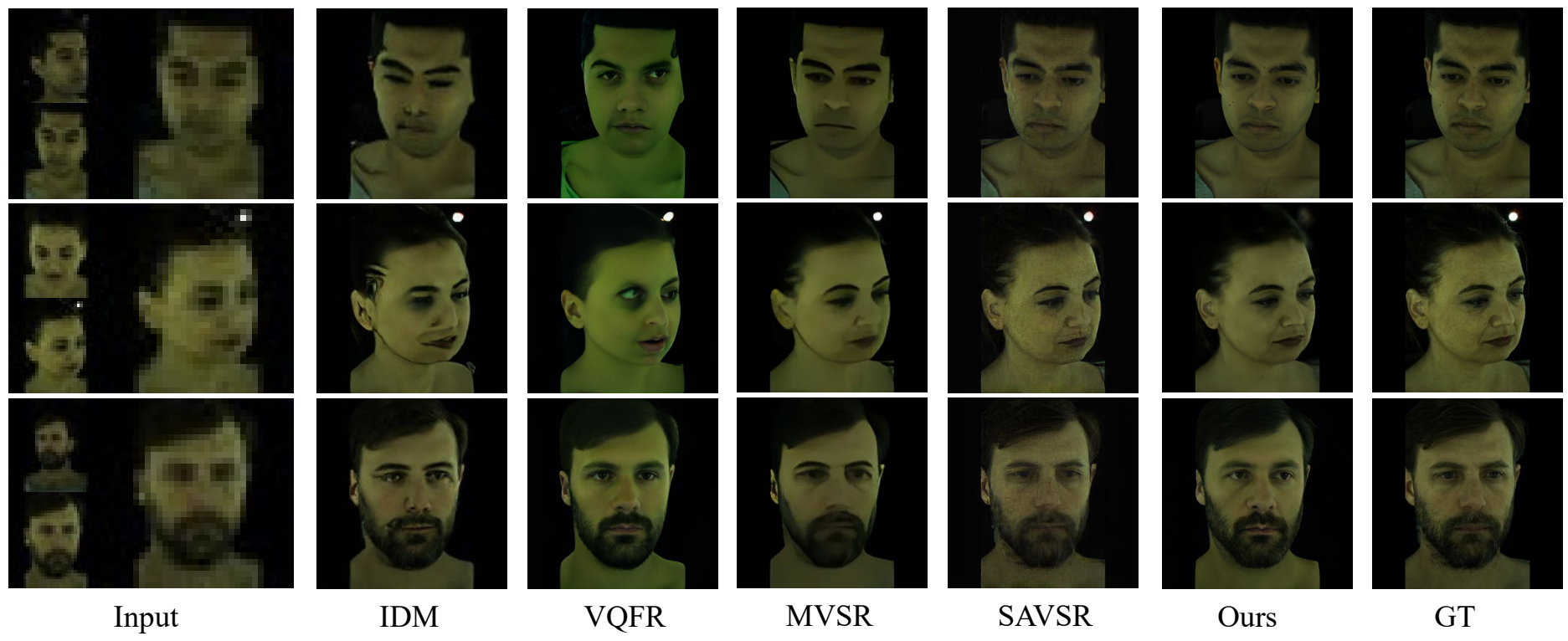

**Fig. 5**: Results of 8x SR on the multiface dataset. [49]

#### 4.3.4 Super-resolution of Highly Degraded Images

We conducted a qualitative comparison experiment of face SR by extracting facial regions from MARS images. For face regions extracted from the MARS dataset, they have sizes around $32 \times 32$, similar to the low-resolution face images that we used in our experiments. Most of the face images extracted from the MARS dataset are highly degraded. The SR results of SwinIR, VQFR, LIIF, IDM and the proposed method are visualized in Fig. 6.

Due to the low quality of the input faces, the SR results tend to be less quality than the results in other experiments. It can be seen that LIIF and IDM improve the details of the input faces, but the results are still largely blurry. SwinIR recovers more facial details than LIIF and IDM, but it can be seen that the eyes, nose, and mouth are not accurately represented, and facial areas, such as the chin, are not clearly distinguishable. VQFR produces a face image with finer details than other methods, but it can be seen that VQFR does not take into account the characteristics of the target, such as identity and gender. It can be seen that the proposed method accurately identifies the characteristics of the target from multiple images and recovers the facial image with finer details in a cascade manner.

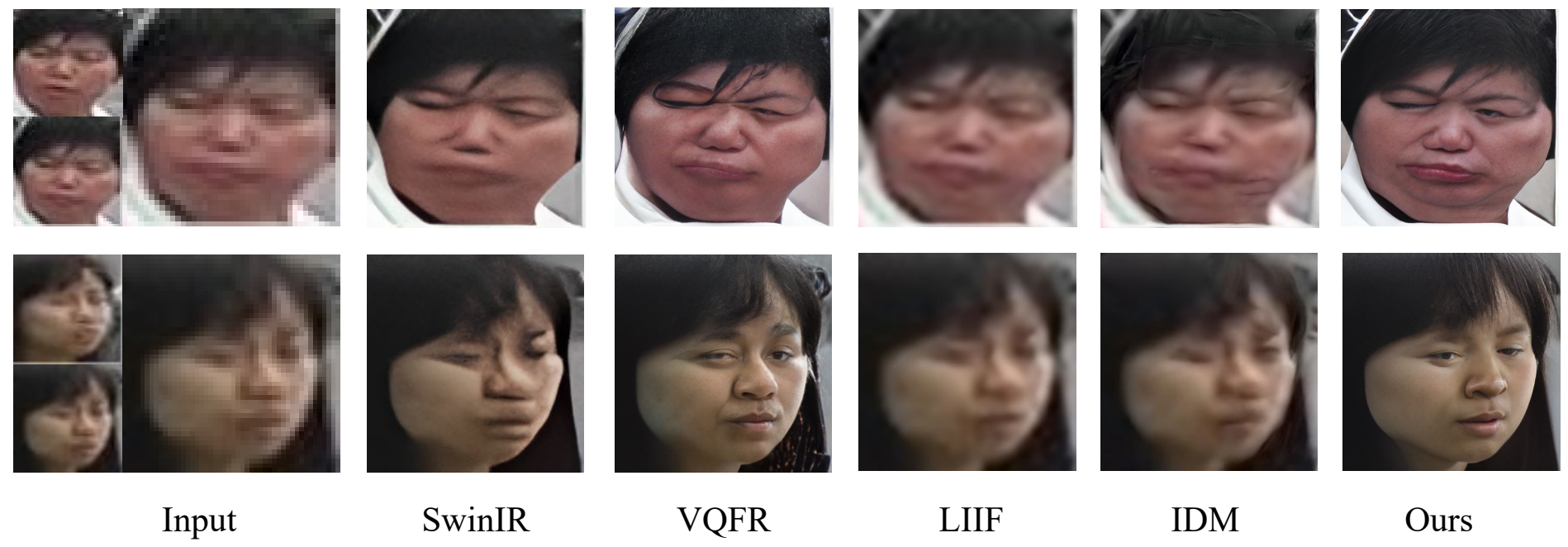

**Fig. 6**: Results of 8x SR on the MARS dataset [10]. We manually crop the face region of the target person. It can be seen that our results accurately restore the details of the image for highly degraded facial images compared to other methods.

### 4.3.5 Person Re-identification

Although the earlier experiments focus on facial image restoration, we additionally conduct an experiment on pedestrian re-identification to evaluate the general applicability of our method beyond facial regions. In this experiment, we apply our super-resolution model to full-body pedestrian images containing a wide range of appearances, poses, and background clutter. To ensure generality, we replace the face recognition module with a ResNet backbone pretrained on ImageNet, enabling identity embeddings without relying on any face-specific priors. This setting allows us to test whether our identity-guided SR approach is still effective in enhancing identity discriminability in general surveillance contexts.

To further verify our method in general and application usages beyond facial superrsolution, we compared our method in re-identification scenarios. Therefore, we add FC layers to the encoder side of the cascade restoration model to regress latent embeddings for person re-identification. To ensure generality beyond face-specific domains, we replaced the face recognition network with a ResNet backbone pre-trained on ImageNet classification tasks[61]. We use the motion analysis and re-identification dataset (MARS) [10] to train our model for video-based re-identification. We first train SR tasks from 128 $\times$ 64 resolution to 256 $\times$ 128 by resizing images on the dataset. Then, we use latent embeddings of the cascade restoration model mentioned above and train the network in an end-to-end manneer for classification. We use cross-entropy classification and triplet loss for training and fine-tuning the classification network.

To ensure a comprehensive and fair evaluation, we compared our method with six state-of-the-art video-based re-identification approaches, including three established baselines [37, 38, 62] and three recent competitive methods, SEAS [63], BAU [64], and our own. These methods were fine-tuned on the MARS dataset resized to 128 $\times$ 64 resolution. Briefly, STMN [37] utilizes temporal memory networks to capture consistent attention across frames. PSTA [38] introduces a spatial-temporal pyramid attention mechanism for robust feature alignment. SINet [62] applies a salient-to-broad transition module to gradually expand attention regions over frames. SEAS [63] leverages shape-aligned supervision to enhance spatial consistency across frames. BAU [64] balances alignment and uniformity in feature space to improve generalization. Our method jointly performs face super-resolution and identity embedding using collaborative identity reconstruction. Following the recent related works [37, 38, 62], we used the maximum a posteriori (mAP) and rank-1 accuracy to measure performance. The quantitative performance results for face re-identification on the MARS dataset are summarized in Table 3.

**Table 3**: Quantitative comparisons on MARS dataset [10] in terms of mAP (%) and rank-1 accuracy (%).

| | STMN [37] | PSTA [38] | SINet [62] | SEAS [63] | BAU [64] | Ours |
|---|---|---|---|---|---|---|
| mAP (↑) | 79.2 | 80.8 | 82.5 | 83.2 | 83.6 | **83.9** |
| rank-1 (↑) | 84.7 | 85.5 | 88.1 | 88.7 | 89.1 | **90.3** |

Our method shows performance improvements of 5.93% in mAP and 6.61% in rank-1 accuracy over STMN [37]. While STMN captures temporal consistency through memory-based attention, it lacks an explicit mechanism to integrate identity information across frames, limiting its ability to distinguish similar-looking individuals in the sequences. Compared to PSTA [38], our method improves mAP by 3.84% and rank-1 by 5.61%. Although PSTA blends spatial-temporal features effectively, it relies on local feature alignment between frames. In contrast, our method directly aggregates identity features, enabling more consistent identity embedding across large pose or appearance variations. SINet [62] achieves competitive results through attention expansion, but still shows a 1.67% gap in mAP and 2.50% in rank-1 compared to ours. This is likely due to its focus on spatial coverage rather than explicit identity preservation. Our method's collaborative identity unification ensures better identity consistency across frames. SEAS [63] improves spatial alignment using shape cues, but does not utilize cross-frame identity fusion. As a result, it underperforms by 0.7% in rank-1 accuracy, especially in sequences with large pose variation or occlusion, where identity-level aggregation provides stronger supervision. Although BAU [64] shows strong generalization via contrastive learning, its per-frame feature encoding lacks the hierarchical reconstruction and identity unification our framework offers. This limits its discriminative power under severe degradation or low-resolution conditions, leading to lower re-identification accuracy than our method. These comparative results demonstrate that beyond robust visual encoding, explicitly integrating identity information across frames, as done in our collaborative reconstruction framework, is crucial to achieving reliable performance in real-world re-identification scenarios.

#### 4.3.6 Super-resolution in Re-identification

We extended our experiments on person re-identification in Sec. 4.3.5 for additional performance analysis. In the experiments in Sec. 4.3.5, we resized the MARS dataset [10] by 0.5x to jointly learn re-identification and 2x SR tasks. Therefore, we validate the performance of the proposed method through SR results in the person re-identification domain. The SR results are illustrated in Fig. 7.

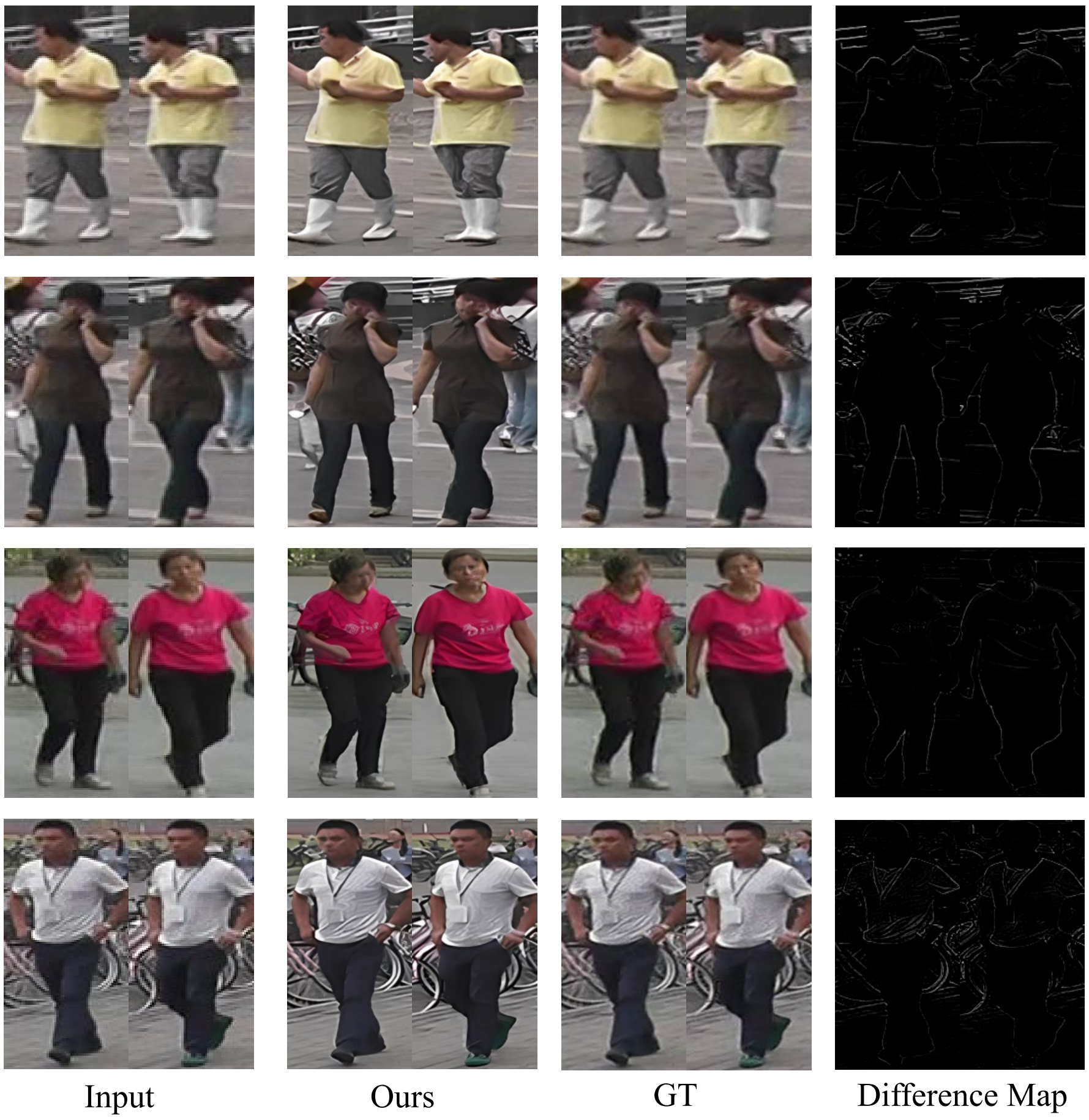

**Fig. 7**: Results of 2x SR on the MARS dataset [10] with the error map. Our method performs re-identification based on SR. The results demonstrate the capability of the proposed method to estimate common features from multiple images.

Our method uses common features extracted from multiple images to collaboratively perform SR. It can be seen that the utilization of common features makes it possible to accurately restore details even in highly degraded images in MARS. Even better, it can be shown that our method restores images that are clearer and less blurred at the edges of objects and people than the ground-truth images. In detail, the error map confirms that the proposed method clearly restores high-frequency details, such as edges on objects and people or wrinkles on clothes.

## 4.4 Ablation Studies

For further evaluation, we conduct qualitative and quantitative evaluations for ablation studies. We evaluated the performance of our model by excluding the collaborative identity network and the cascade architecture, respectively. We show the quantitative comparisons of the ablation studies in Table 4.

**Table 4**: Quantitative results for ablation studies.

| | w/o unified ID | w/o cascade | Ours |
|---|---|---|---|
| PSNR (↑) | 22.82 | 22.04 | 25.58 |
| SSIM (↑) | 0.6143 | 0.6058 | 0.6977 |
| LPIPS (↓) | 0.4314 | 0.4557 | 0.3458 |
| ID (↑) | 0.674 | 0.647 | 0.792 |
| FID (↓) | 55.21 | 56.14 | 51.42 |

The overall quantitative quality shows that the network without the cascade architecture performs the worst, and the network without the collaborative identity network performs better than the network without cascade architecture. The results show that our cascade approach in the restoration network allows the SR network to accurately restore the high-quality facial image and identity through multiple cascade iterations. Finally, these results demonstrate that the unified facial identity obtained through our collaborative identity network plays an important role in restoring facial detail in our framework.

Fig. 8 shows qualitative comparisons of the $8\times$ face SR. Without the collaborative identity, the facial identity with insufficient target information is extracted from an input image. Therefore, the restoration results without the collaborative identity are shown to be inaccurate, since facial features extracted from a single image do not have sufficient information for the target face. As a result, the restored facial images are shown to lack details such as facial expressions, eye size, and lip shape, compared to those with the collaborative identity. Without a cascade architecture, the restoration task of the facial image and its identity cannot be performed progressively. Thus, the method without a cascade architecture performs the SR task with the initial image features and facial identity extracted from the low-resolution input image. This causes some artifacts in the SR images, such as an unnatural gaze direction, as shown in Fig. 8.

Furthermore, to examine the appropriate number of faces used for the collaborative identity reconstruction, we measure PSNR and ID scores of PSNR and ID scores by varying the number of images from 1 to 5. Fig. 9 depicts the performance measurements according to the number of facial images. PSNR and ID scores of the network using a single image show the lowest performance. When using two or more images, the results show no significant differences from each other. The results show that collaborative identity reconstruction with a pair of images is sufficient to find the common features of an individual. Throughout our experiments, we set $N = 2$ to balance performance and computational cost. As shown in Fig. 9, increasing $N$ beyond 2 yields marginal improvements, indicating that a pair of images is sufficient to construct a robust identity representation for most cases. This setting was empirically validated and used consistently for fair comparisons.

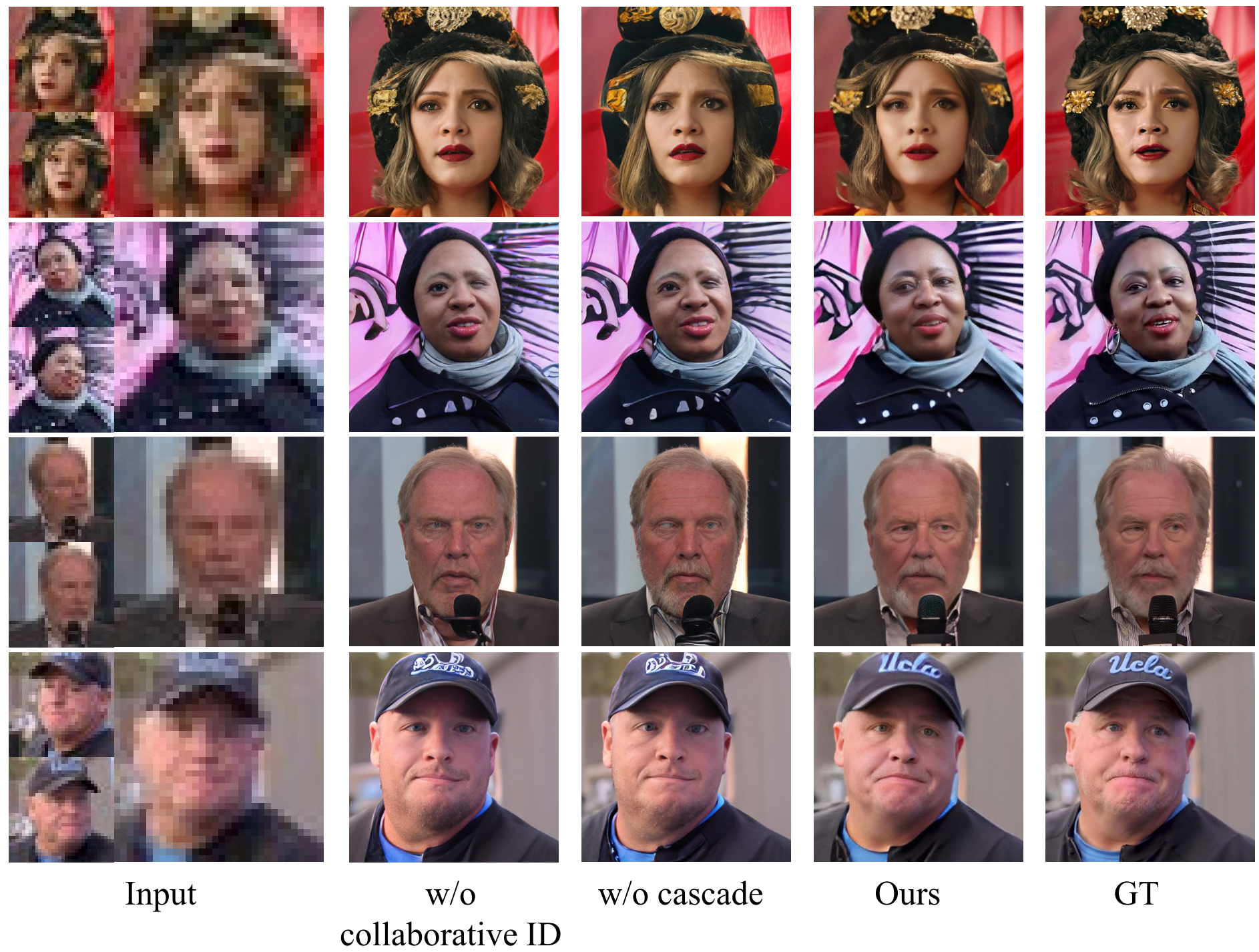

**Fig. 8**: Comparison of qualitative results for ablation studies.

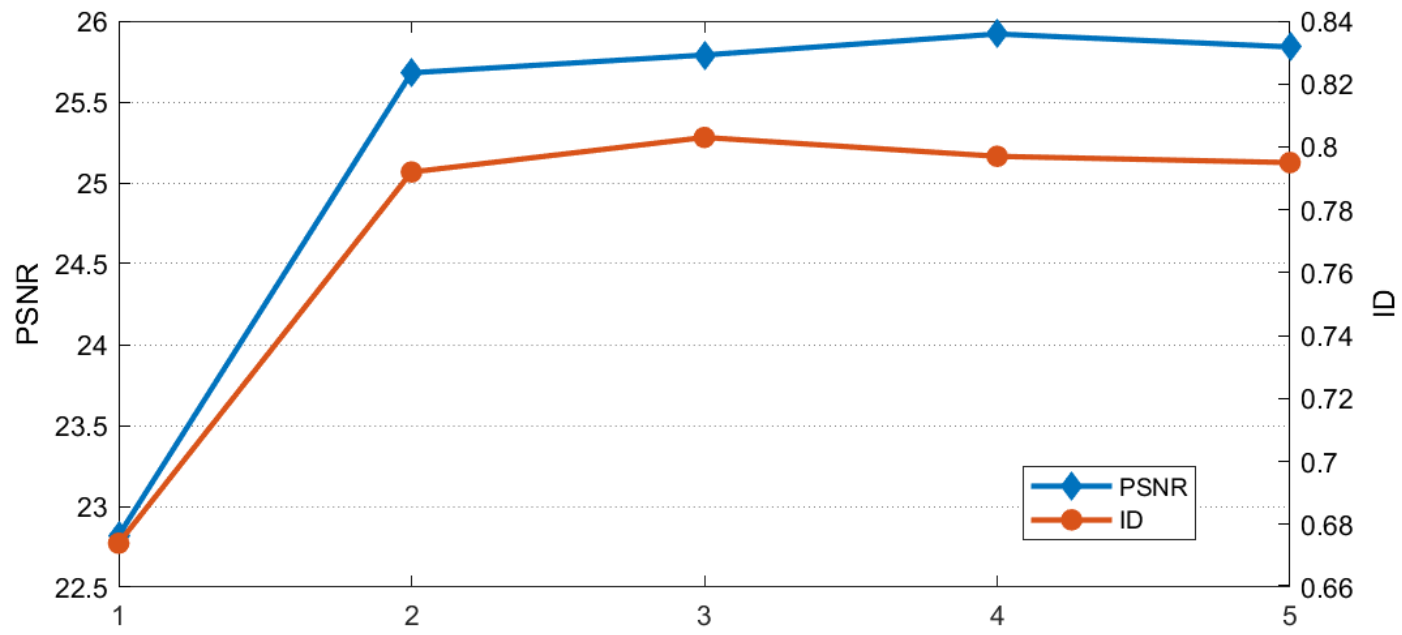

**Fig. 9**: Comparison of the performance changes according to the number of images used for the collaborative identity network.

In addition, to assess the dependency of our collaborative face SR framework on the choice of identity feature network, we conducted a controlled experiment comparing three widely used face recognition backbones: ArcFace [41], AdaFace [52], and TopoFR [53]. These models were selected for their popularity and performance across different face recognition settings. ArcFace, which we used as the default in our main experiments, employs an additive angular margin loss to enhance class separability in the embedding space. AdaFace dynamically adjusts the classification margin based on input quality, making it more robust to low-quality faces. TopoFR focuses on preserving topological structure between identity embeddings by aligning relational graphs during training.

We replaced the identity feature network $\phi(\cdot)$ in our collaborative identity module with each of the three backbones while keeping all other network components fixed. For fair evaluation, we followed the same experimental setup as Section 4.3.1, using the VFHQ [47] dataset and evaluating with five metrics: PSNR, SSIM, LPIPS, identity similarity (ID), and FID. This setting was chosen because it highlights face-specific reconstruction quality under real-world temporal variations, where identity preservation is crucial.

**Table 5**: Comparison of different identity feature networks in our framework. Experiments are conducted on VFHQ dataset under the same 8× SR setting as Section 4.3.1.

| Backbone | PSNR (↑) | SSIM (↑) | LPIPS (↓) | ID (↑) | FID (↓) |
|---|---|---|---|---|---|
| ArcFace [41] | 25.58 | 0.6977 | 0.3458 | 0.792 | 51.42 |
| AdaFace [52] | 25.41 | 0.6913 | 0.3512 | 0.785 | 52.04 |
| TopoFR [53] | 25.76 | 0.7015 | 0.3427 | 0.796 | 50.87 |

From Table 5, we observe that all three identity feature networks lead to similar SR performance, with PSNR differences within 0.35 dB and LPIPS/ID variations below 0.01. TopoFR achieved slightly better results across all metrics, likely due to its emphasis on relational feature structure, which may benefit multi-image identity fusion. AdaFace performed marginally lower, which may be attributed to its quality-adaptive design being more effective in variable-quality datasets than in preprocessed video sequences.

Importantly, these results confirm that our SR framework is robust to the choice of identity network. The backbone does not significantly alter performance, suggesting that the proposed collaborative identity aggregation and cascade refinement architecture are the primary contributors to the system's effectiveness, rather than any specific recognition model. This also supports the generality of our method, as it can integrate various identity embedding models without significant re-tuning or performance degradation.

# 5 Conclusions

We have introduced a novel face super-resolution method with collaborative feature reconstruction from multiple facial images of an individual. Our framework uses common facial features in multiple images to achieve SR tasks based on additional facial information that can hardly be obtained from an individual low-resolution image alone. Futhermore, we have performed facial identity fusion through a transformer-based collaborative identity network to estimate the common facial identity features from multiple images for the cascade SR framework. We have accomplished SR tasks stably from sequential or multi-view images by progressively restoring facial images with facial features obtained from the collaborative identity network. Our key innovations include a cascade SR network that performs progressive restoration via residual learning, and a transformer-based collaborative identity aggregation module that fuses identity features from multiple views into a unified identity prior. These components enable accurate recovery of high-frequency details and robust identity preservation. The experimental results and comparisons have shown that our framework achieves significant improvements over other state-of-the-art methods, especially in face SR. Moreover, our method has shown comparable performance to other state-of-the-art methods even when used for person re-identification, demonstrating general performance of the proposed method. Nevertheless, our approach has certain limitations. In particular, the multi-stage cascade structure introduces additional computational overhead, and the reliance on multiple input images may limit its applicability in scenarios where only a single low-resolution image is available. Additionally, while our collaborative identity aggregation effectively unifies facial features, its performance may degrade under extreme pose variations or occlusions. Future work will focus on optimizing the cascade network for computational efficiency and exploring adaptive aggregation techniques to handle a wider range of input conditions, including single-image scenarios. Moreover, integrating more advanced attention mechanisms and incorporating multi-modal cues may further enhance the robustness and generalizability of the proposed method to other collaborative applications.

# Declarations

## Author contribution

Juheon Hwang: Investigation, Software, Writing - original draft, Taewan Kim: Writing – review & editing, Jiwoo Kang: Conceptualization, Methodology, Supervision, Writing – review & editing.

## Funding

This work was supported by Institute of Information & communications Technology Planning & Evaluation (IITP) grant funded by the Korea government (MSIT) (No. RS-2023-00229451, Interoperable Digital Human (Avatar) Interlocking Technology Between Heterogeneous Platforms).

## Data availability

We used the VFHQ [47] and CelebV-HQ [48] datasets for sequential supperresolution. Also, we used the Multiface dataset [49] for multiview supperresolution. These dataset are publicly accessible online: VFHQ in https://liangbinxie.github.io/projects/vfhq, CelebV-HQ in https://github.com/CelebV-HQ/CelebV-HQ, and Multiface dataset in https://github.com/facebookresearch/multiface.

## Conflict of interest

The authors declare no competing interests.

## Ethics approval

Not applicable.